# When AI Does Science: Evaluating the Autonomous AI Scientist KOSMOS in Radiation Biology

Humza Nusrat[1,2], Omar Nusrat[3]

1) Henry Ford Health, Detroit, USA; 2) Michigan State University, East Lansing, USA, 3) Toronto Metropolitan University, Toronto, Canada

## 1. Abstract

Recent advances have led to the development of agentic AI scientists. These complex systems can perform literature searches, computational processing, and hypothesis generation. In this work, we evaluate KOSMOS, an autonomous AI scientist, on three complex problems in radiobiology. For each proposed hypothesis, the analysis performed by KOSMOS was reconstructed and each claim audited against simple empirical null models based on random gene sets or random signatures. Hypothesis 1: baseline DNA damage response (DDR) capacity across cell lines would predict the p53 transcriptional response after irradiation (GEO dataset: GSE30240). Hypothesis 2: baseline expression of OGT and CDO1 would predict the strength of repressed and induced radiation response modules in breast cancer cells (GEO dataset: GSE59732). Hypothesis 3: a 12-gene expression signature as a predictor of biochemical recurrence-free survival after prostate radiotherapy plus androgen deprivation therapy (GEO dataset: GSE116918). The DDR to p53 hypothesis was not supported by our analysis. Across four cell lines, the DDR score and p53 response were weakly negatively correlated (Spearman $\rho = -0.40$, $p = 0.756$). This was indistinguishable from five-gene random scores. The OGT and CDO1 hypothesis yielded mixed results. OGT showed only a weak correlation with the repressed module ($r = 0.23$, $p = 0.341$), whereas CDO1 was a clear outlier as a predictor of the induced module ($r = 0.70$, empirical $p = 0.0039$). Finally, the 12-gene signature achieved a concordance index of 0.613, outperforming random signatures ($p = 0.0166$). However, its hazard ratio was not unusually large ($|\log HR| = 0.899$, $p = 0.3738$), indicating that it does not capture a prognostic signal. In this case study, KOSMOS delivered one discovery that was well supported (CDO1), one plausible but uncertain result (the 12-gene signature), and one false result (DDR to p53). The work demonstrates that AI systems can generate useful hypotheses in radiation biology, however, their outputs must be examined with rigor and tested against appropriate null models.

**Key words:** Large language model (LLM), AI scientist, autonomous research agent, agentic LLM, scientific discovery, trustworthy AI, radiation biology

## 2. Introduction

Recently, several agentic AI scientists have been proposed.[1–4] These systems have yet to undergo independent, scientific evaluation. The core reasoning component of recently developed autonomous AI scientists are large language models (LLMs). LLMs are prone to hallucination and can propose numerous hypotheses using convincing, well-written language.[5] Without rigorous evaluation, fluency can seem like thoughtful reasoning. In this work, we argue that evaluating an AI system's potential as a 'scientist' requires testing its claims using quantitative evaluation. Towards this, we propose a falsification-based auditing methodology based on Popper's principle of falsification.[6] Each AI claim must be testable and falsifiable.

In this work, we evaluate the recently released AI scientist, KOSMOS.[1] We treat KOSMOS's outputs as preliminary hypotheses rather than accepted results. By formulating appropriate null models and empirical benchmarks for each claim, we ensure that any purported insight is considered valid only if it proves statistically significant. Simple, domain-appropriate null model experiments were used to quantitatively evaluate KOSMOS's reported discoveries.

We quantitatively evaluated three hypotheses generated by KOSMOS in radiation biology. The radiation biology datasets used for evaluation in this work span both cellular and clinical domains.

1. **Cell-line DNA damage response (GSE30240[7])**: KOSMOS hypothesized that a cell line's baseline DNA damage response (DDR) capacity predicts the magnitude of its p53 transcriptional response to ionizing radiation. We tested this crossline correlation using null simulations.
2. **Radiation module regulators (GSE59732[8])**: KOSMOS identified two genes, OGT and CDO1, whose basal expression was proposed to quantitatively predict the strength of radiation-responsive gene modules in breast cancer cells. We audited each candidate regulator's effect using random gene benchmarks.
3. **Prostate survival signature (GSE116918[9])**: KOSMOS proposed a 12-gene expression signature as a predictor of biochemical recurrence-free survival in prostate cancer patients receiving radiotherapy plus androgen deprivation therapy (ADT). We evaluated the signature's prognostic power against thousands of random 12-gene signatures.

Through these analyses, we seek to answer a fundamental question: can an AI scientist not only generate hypotheses but also contribute true scientific insights? We report one clear success, one failure, and one ambiguous outcome. These results illustrate both the promise and the pitfalls of relying on AI for hypothesis generation. These findings demonstrate that AI scientists (such as KOSMOS) are promising tools for the advancement of science. However, in their current state, they should be viewed as powerful tools that require auditing and quality assurance instead of completely autonomous.

## 3. Methods

### 3.1. Overall auditing strategy

For each KOSMOS hypothesis, a minimal statistical test was designed along with an empirical null distribution. Simple correlation and survival analyses were used to examine proposed effects. Each claim's empirical p-value was calculated using its observed test statistic and a null distribution (generated from many random permutations or random gene sets).

#### 3.1.1. GSE30240 (baseline DDR vs p53 response)

For each of the four human cell lines in GSE30240[7] with complete data (BJ, G361, HepG2 and TK6), "DDR competence" score was calculated and used as the mean baseline (0 Gy) expression of the five DNA damage response genes specified by KOSMOS. For each line, we also computed a "p53 response". This response was defined as the average induction (log2 fold change) of nine p53 target genes (at 6 hours after irradiation). The association between the DDR score and the p53 response was then assessed across cell lines using Spearman rank correlation (non-parametric given N = 4).[10] To build a null distribution, 10,000 random five-gene "DDR gene sets" of the same size as the true set were generated. For each random set, we recalculated the baseline DDR score for each cell line and correlated that score with the p53 response values (yielding 10,000 Spearman ρ values representing random expectations under the assumption of no genuine DDR to p53 link). We then computed a two-sided empirical p-value as the proportion of random gene sets whose |ρ| was at least as large as the observed |ρ| from the KOSMOS DDR gene set.

#### 3.1.2. GSE59732 (OGT/CDO1 and radiation module strength)

In a panel of 16 breast cancer cell lines [8] (0 Gy vs 5 Gy, 24 h), KOSMOS hypothesized that two genes, OGT[11] and CDO1[12], act as upstream regulators which predict the magnitude of radiation-responsive expression modules. In order to evaluate this, radiation-response modules were derived directly from the dataset. For each line, the per-gene log2 fold change (5 Gy vs 0 Gy) was computed and the 250 genes with the highest variance across lines were selected. They were then clustered into three groups (based on correlation distance and average linkage). The cluster with the most negative mean fold change was classified as "Cluster 1" (repressed module) and the cluster with the most positive mean fold change as "Cluster 3" (induced module). For each cell line, a module response score was computed as the mean log2 fold change of genes in each cluster. For both OGT and CDO1, baseline expression levels (0 Gy) across the 16 lines were extracted. We then tested if the baseline expression of each gene correlated with the module response scores across the 16 lines using Pearson correlation. To determine significance, we compared each candidate gene's correlation against a null distribution based on 10,000 random genes from the same expression matrix (excluding OGT and CDO1). For each random gene, we computed its correlation with the module response score. We then computed a two-sided empirical p-value as the fraction of random genes whose |r| was as large as or larger than the observed |r| for the candidate gene.

### 3.1.3. GSE116918 (12-gene survival signature in prostate cancer)

KOSMOS proposed a 12-gene expression signature for classifying patients by prognosis (biochemical recurrence-free survival) after RT plus ADT for prostate cancer. GSE116918[9] contains gene expression profiles from primary prostate tumors and the associated times to biochemical recurrence after RT + ADT. We reconstructed KOSMOS's 12-gene risk score by computing the mean expression of the 12 specified genes (for each patient). This was used as the covariate in a univariate Cox proportional hazards model in order to predict the time to biochemical recurrence.[13] We quantified prognostic performance in two ways. Firstly, we measured discrimination using the concordance index (c-index). Secondly, we measured effect size using the Cox log hazard ratio (logHR) associated with the signature score. To test whether the observed performance was genuinely special, we generated 5,000 random 12-gene signatures by repeatedly sampling 12 genes from the same expression dataset (excluding the original signature genes). For each random signature, we computed an analogous mean expression score per patient and fit a univariate Cox model. This procedure yielded null distributions for c-indices and |logHR| values. For each metric, we computed an empirical p-value as the proportion of random signatures with c-index greater than or equal to the observed c-index, and with |logHR| greater than (or equal to) the observed |logHR|.[14]

### 3.2. Reproducibility

All datasets used in this study are publicly available, with GEO accession numbers provided in the text. The analysis code for computing scores, correlations, Cox models and null distributions was written in Python and R and will be made available upon request. Randomization procedures were performed with fixed random seeds. All KOSMOS prompts used to generate the hypotheses, as well as corresponding KOSMOS artifacts reports have been made publicly available.[15]

## 4. Results

### 4.1. Hypothesis 1: baseline DDR capacity vs p53 response amplitude (GSE30240)

There was a weak negative correlation between DDR competence score and p53 response score (Spearman ρ = -0.40) across the four complete cell lines (BJ, G361, HepG2 and TK6). Lines with higher baseline DDR gene expression exhibited slightly lower p53 induction. This correlation was not statistically significant given the small sample size (n = 4 lines; p approximately 0.60 by Spearman test). Null simulations confirmed that such a correlation was typical of random gene sets. The null distribution of Spearman ρ values from 10,000 random five-gene "DDR-like" sets was centered near zero. 75.6% of random five-gene sets yielded a correlation of at least 0.40 (empirical p = 0.756). The proposed DDR gene set did not outperform any given random five-gene score from the dataset. Hypothesis 1 was refuted under our analysis. A scatterplot of the observed data (DDR score vs p53 response) with the null expectation (Figure 1) demonstrates the small-sample noise and the lack of deviation from random behavior.

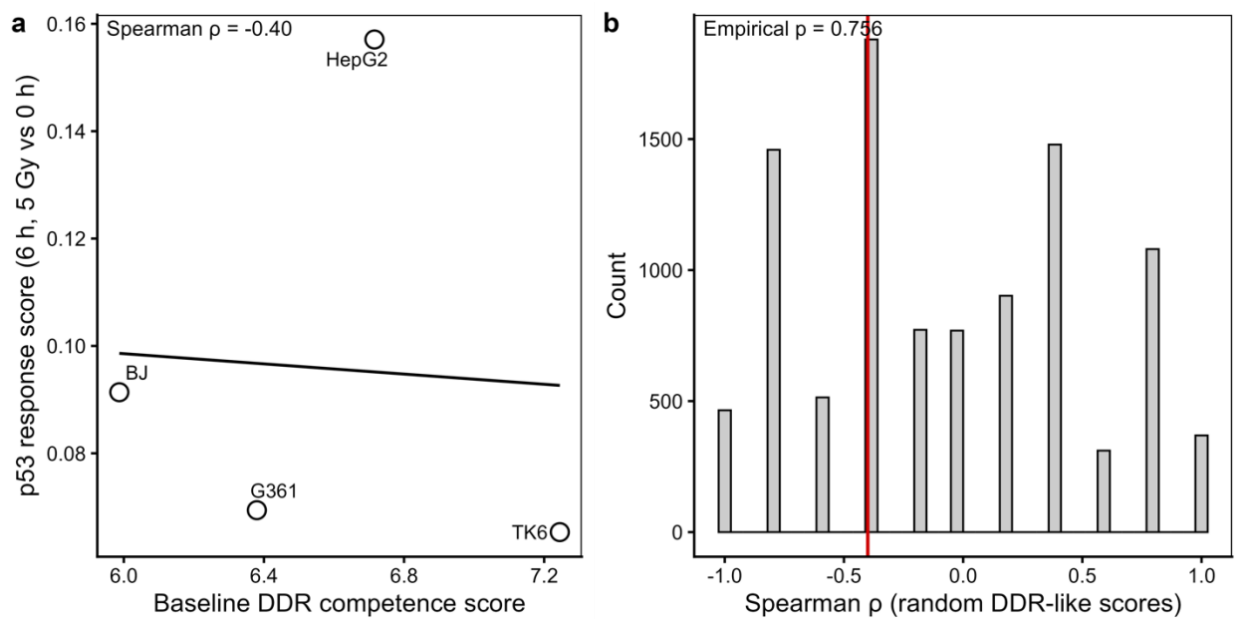

Figure 1: Baseline DDR vs p53 response data and null benchmark (GSE30240): (a) Scatter plot of the four cell lines showing p53 response score (y-axis; mean log2 fold change of p53 targets at 6 hours, 5 Gy vs 0 hour, 0 Gy) versus baseline DDR competence score (x-axis; mean 0 Gy expression of the five KOSMOS DDR genes), with a fitted trend line. (b) Histogram of Spearman correlation coefficients obtained from 10,000 random five-gene "DDR-like" sets (gray bars). The red vertical line marks the observed ρ, which lies well within the null distribution, with 75.6% of random sets achieving an equal or larger |ρ| (empirical p = 0.756).

## 4.2. Hypothesis 2: baseline OGT and CDO1 expression vs radiation module strength (GSE59732)

Hypothesis 2 evaluated the two candidate upstream regulators that KOSMOS predicted for radiation response heterogeneity in breast cancer cell lines.

### 4.2.1. OGT and Cluster 1 repression

KOSMOS reported that baseline expression of OGT (O-GlcNAc transferase) correlated with the magnitude of a Basal-enriched repression module (Cluster 1 genes) across lines. In our re-analysis, OGT showed only a weak positive correlation with the Cluster 1 response score (Pearson $r = 0.23$). When compared against random genes, this correlation was found to be expected. In the empirical null distribution of correlations between random genes and the Cluster 1 score, 34.1% of genes achieved $|r|$ greater than or equal to 0.23 (empirical $p = 0.341$). Therefore, OGT did not meet our significance cutoff ($p < 0.05$).

### 4.2.2. CDO1 and Cluster 3 induction

CDO1 (cysteine dioxygenase) was proposed by KOSMOS to predict the strength of an induced gene module (Cluster 3) enriched in Basal-like lines. Our analysis strongly supported this claim. CDO1 baseline expression had a clear positive correlation with the Cluster 3 induction score (Pearson $r = 0.70$). In the random-gene null test, CDO1 was a marked outlier: only 0.39% of 10,000 random genes attained an equal or larger $|r|$ (empirical $p = 0.0039$). Figure 2 illustrates these results with correlation plots and null distributions for both genes.

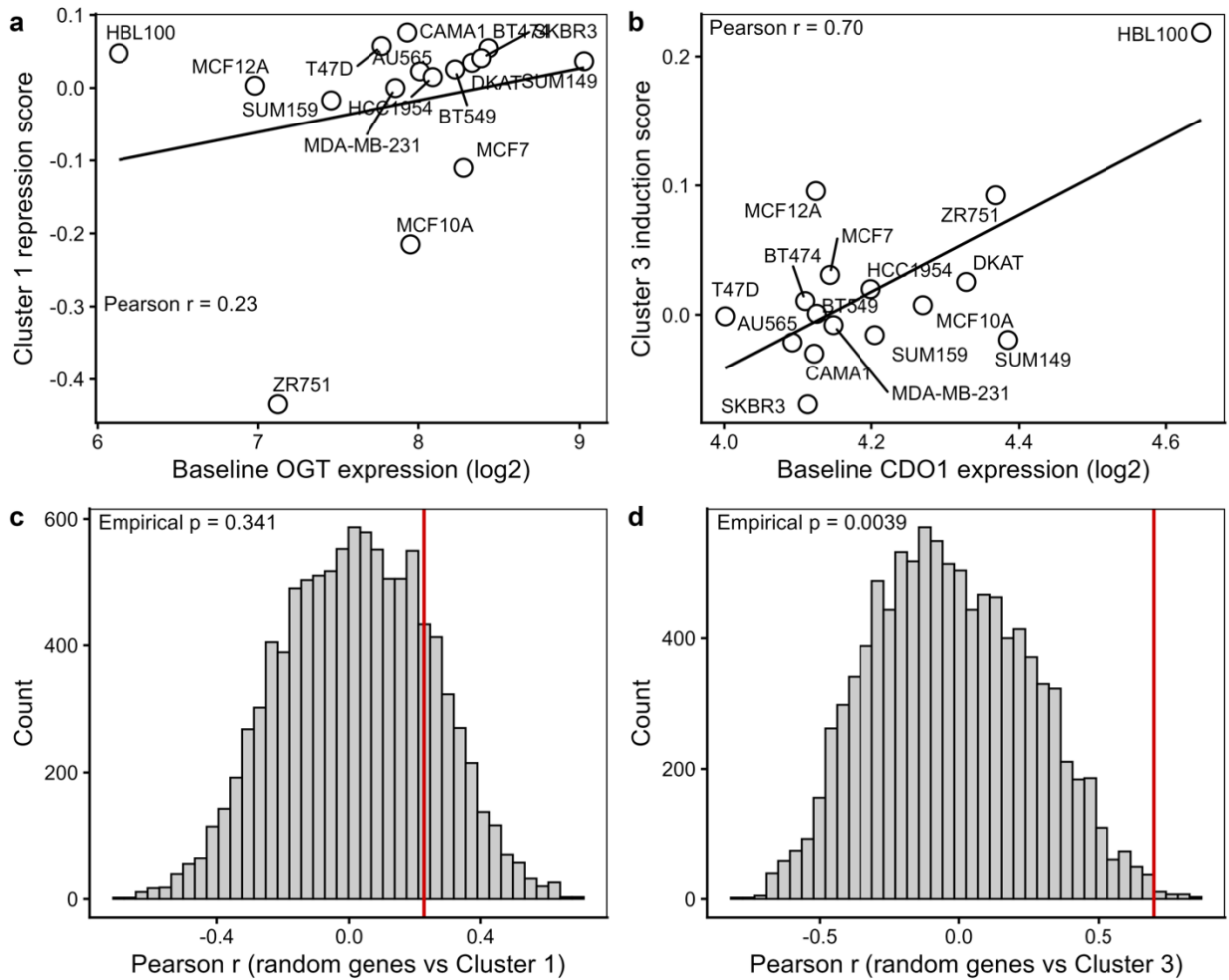

Figure 2: Testing KOSMOS's gene regulators: OGT vs CDO1 (GSE59732): (a) Scatter plot of baseline OGT expression (x-axis, log2) versus Cluster 1 repression score (y-axis; mean log2 fold change of Cluster 1 genes after 5 Gy) across 16 breast cancer cell lines. (b) Scatter plot of baseline CDO1 expression versus Cluster 3 induction score. (c) Null distribution of Pearson r for correlations between 10,000 random genes and the Cluster 1 repression score (gray histogram). The red vertical line marks OGT's observed r, which lies near the center of the null distribution (empirical p = 0.341). (d) Null distribution of Pearson r for correlations between 10,000 random genes and the Cluster 3 induction score. The red vertical line marks CDO1's r, which falls in the extreme right tail (empirical p = 0.0039).

## 4.3. Hypothesis 3: KOSMOS 12-gene signature for prostate cancer survival (GSE116918)

Hypothesis 3 examined the prognostic power of the AI-proposed 12-gene expression signature in a clinical prostate cancer cohort. In 248 patients with complete data, the KOSMOS 12-gene risk score achieved a concordance index of 0.613 in a univariate Cox model for biochemical recurrence-free survival. This c-index score represents a modest discrimination (random guessing being 0.5). The corresponding Cox coefficient yielded a log hazard ratio of 0.899 (hazard ratio = 2.46, Wald p = 0.007), indicating that higher signature scores were associated with increased recurrence.

Figure 3a shows the distribution of c-indices for 5,000 random 12-gene signatures. The vast majority clustered around 0.5 (no predictive power), with the upper tail extending into the 0.6 to 0.7 range. The proposed signature's c-index of 0.613 lay near the top of this distribution with an empirical p = 0.0166. Figure 3b depicts the null distribution of |logHR| for random signatures. Many random gene sets produced hazard ratios in the same range as the proposed signature, which had |logHR| = 0.899. The proposed gene combination did not produce an unusually large separation in survival risk between patient groups, leading to this result being classified as ambiguous. The AI-generated signature demonstrated genuine signal, with a c-index higher than most random sets and thus some non-random prognostic power. It failed to demonstrate uniqueness in terms of effect size. Many random gene sets with enough proliferative genes would be expected to similarly predict outcome, a phenomenon documented in prior studies.[16]

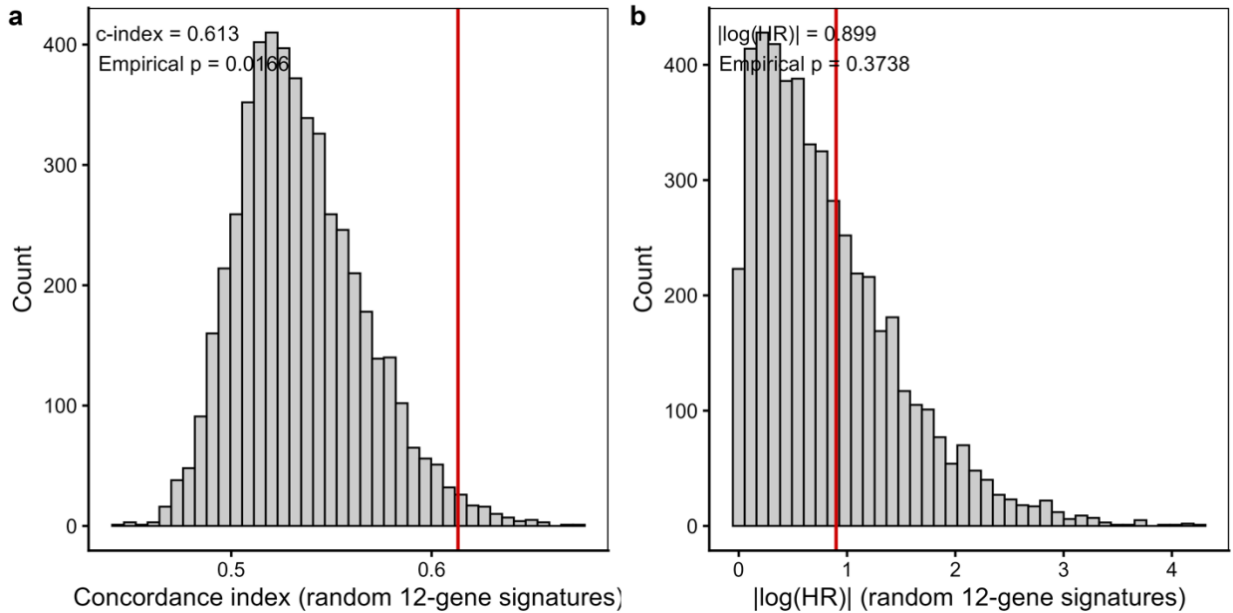

**Figure 3: Evaluating the 12-gene prognostic signature vs random expectations (GSE116918):** (a) Null distribution of concordance indices from 5,000 random 12-gene signatures (gray histogram) in the prostate cancer cohort. The c-index of proposed signature (red line) is 0.613, placing it in the top approximately 2% of random sets (empirical p = 0.0166). (b) Null distribution of the absolute log hazard ratio |logHR| for the same 5,000 random signatures. The proposed signature's |logHR| is 0.899 (red line), which does not fall in the extreme tail of this null (empirical p = 0.3738), suggesting that its effect size (hazard separation) is not unusually large compared with many random gene combinations.

## 5. Discussion

KOSMOS demonstrated an ability to generate biologically sensible hypotheses autonomously. The success with CDO1 in Hypothesis 2 shows how an AI scientist can find a non-obvious regulator that remains significant under statistical testing. CDO1 is not a classic "usual suspect" in radiation response, yet KOSMOS's analysis nominated it, and our validation confirms that it is a strong predictor of transcriptional response variation. This result suggests that AI systems can augment human researchers by flagging novel insights that might otherwise be overlooked. The failures in Hypotheses 1 and 3 illustrate that AI-generated ideas can also be misleading. The DDR versus p53 correlation in Hypothesis 1 was presented as an interesting discovery. However, the underlying evidence was weak and did not differ from random expectations. The 12-gene signature in Hypothesis 3 was found to be modestly effective but did not pass our discovery threshold. Our null model tests demonstrated that it may largely capture known signals, such as proliferation, or patterns that random gene sets can reproduce.

KOSMOS, like other LLM-based systems, produces articulate explanations that connect data and literature. This fluency can make hypotheses seem well founded, but it can also mask underlying uncertainty. Our results emphasize that quantitative truth is not determined by how coherent a research paper appears. For example, KOSMOS likely rationalized the DDR–p53 link using known biology, such as the need for upstream signaling to activate p53 and assembled a convincing mechanistic story. Our auditing approach treats AI outputs as hypotheses to falsify. This proved useful in separating valid insights from hallucinations. Null models and random genes were used to set baselines against which KOSMOS results were verified. From this framework, Hypothesis 1 was flagged as false, since the data were indistinguishable from noise. It provided a nuanced evaluation of Hypothesis 3, where there was some prognostic signal but also evidence of non-uniqueness. It also strengthened the credibility of Hypothesis 2 by demonstrating that CDO1 was a clear outlier relative to random genes.

An additional insight is that AI may often find known patterns under new guises. The prostate 12-gene signature in Hypothesis 3 likely correlates with cell proliferation, given that random gene sets enriched for cell cycle genes also predict survival. OGT has links to metabolism and stress responses that could connect to established pathways. In such cases, the AI scientist may be overfitting to an implicit signal that is broadly present in the data. The main risk in such cases seems to be overstatement of novelty. Our null model approach helps reveal when an AI discovery is a repackaging of generic signals. For example, many gene sets can achieve similar performance to the 12-gene signature which tempered uniqueness claims. AI-driven analyses may benefit from explicit checks for overlap with known signatures and from domain-informed null models to distinguish genuinely novel insights from obvious or redundant ones.

In the case of CDO1 in Hypothesis 2, the CDO1 signal was supported by a larger sample size (16 lines), increasing statistical power. The effect size was large and specific. CDO1 stood out among many genes, suggesting that KOSMOS focused on a genuine outlier rather than on a generic trend. This success also reflects the benefit of combining multiple sources of information. KOSMOS did not only examine correlations, but also drew on literature about metabolism and stress, which may have guided it toward CDO1 as a biologically plausible candidate. AI hypotheses that integrate both data patterns and prior knowledge may have a higher chance of being valid. Identifying the common features of such successful AI-generated hypotheses is an interesting direction for future work.

The mixed results across Hypotheses 1, 2 and 3 suggest how an AI scientist might best be used in practice. Rather than attempting to replace human intuition or experimental work, systems like KOSMOS may be most effective as prolific generators of ideas that broaden the hypothesis space. They can connect results from literature and data at a speed and scale that individual researchers cannot match. Each proposed hypothesis then becomes a starting point for the traditional scientific method: design analyses or experiments to test it. The successful hypothesis adds to our knowledge and can be followed up with mechanistic studies. In this case, CDO1 now merits further investigation to understand how its expression modulates radiation response, which may open a new line of research. This kind of synergy can be powerful if it is used carefully. AI systems can generate questions, while human scientists and careful statistical checking are used to verify the answers.

We propose a simple workflow for auditing AI-generated science, consisting of a short checklist:

1. Formulate each AI claim in a testable and falsifiable manner.
2. Select appropriate statistical tests or models that directly address the claim.
3. Construct domain-relevant null distributions, for example using randomization or simulation, to define what "by chance" looks like.
4. Establish domain-relevant thresholds for significance and classify hypotheses as supported, refuted or inconclusive based on their position relative to the null.

Automating parts of this pipeline, as some recent tools aim to do, could be valuable as AI systems begin to produce many candidate hypotheses.[1–4] In the longer term, one can imagine workflows in which AI is treated as a junior collaborator that can brainstorm and even draft analyses, but an accompanying "AI auditor" or human scientist independently verifies the results. Such a structure would reduce the risk of false findings and increase confidence in the discoveries that survive this scrutiny.

We acknowledge several limitations of our study. Our null model tests, while rigorous, do not reflect the complexity of the underlying cell biology evaluated in this work. For example, the random gene set approach implicitly assumes that genes are exchangeable. This is an oversimplification since genes within the same pathway are not independent. We attempted to make the nulls fair by matching gene set sizes and using the same underlying expression matrices, but more refined null models, such as ones that preserve network degree or expression distribution, could be explored. In addition, our confirmation analyses were retrospective and used the same datasets that KOSMOS analyzed. Ideally, AI-generated hypotheses would be tested on independent datasets or validated experimentally. In this work we were limited to in silico tests, which can confirm associations but not causation.

# 6. Conclusion

An autonomous AI scientist (KOSMOS) was evaluated in the domain of radiation biology. The system yielded one well-supported discovery, one plausible but unproven lead, and one false result. As these systems become more prevalent, establishing a standardized AI hypothesis auditing protocol will be increasingly important to preserve the credibility of scientific literature.